\documentclass[graybox]{svmult}

\usepackage{mathptmx}       
\usepackage{helvet}         
\usepackage{courier}        
\usepackage{type1cm}        
\usepackage{makeidx}         
\usepackage{graphicx}        
\usepackage{multicol}        
\usepackage[bottom]{footmisc}

\makeindex             

\begin{document}

\title*{Conversational Agents and Children:\\ Let Children Learn}
\author{Casey Kennington and Jerry Alan Fails and Katherine Landau Wright and Maria Soledad Pera}
\institute{
Casey Kennington \at Computer Science, Boise State University, USA, \email{caseykennington@boisestate.edu}
\and Jerry Alan Fails \at Computer Science, Boise State University, USA, \email{jerryfails@boisestate.edu}
\and Katherine Landau Wright \at Literacy, Language and Culture, Boise State University, USA, \email{katherinewright@boisestate.edu}
\and Maria Soledad Pera \at Web Information Systems, TU Delft, The Netherlands, \email{m.s.pera@tudelft.nl}}
%
%
\maketitle

\abstract{  
Using online information discovery as a case study, in this position paper we discuss the need to design, develop, and deploy (conversational) agents that can -- non-intrusively -- guide children in their quest for online resources rather than simply finding resources for them. We argue that agents should ``let children learn" and should be built to take on a teacher-facilitator function, allowing children to develop their technical and critical thinking abilities as they interact with varied technology in a broad range of use cases.\footnote{This paper was accepted to the Language-Based AI Agent Interaction with Children @IWSDS'23, Los Angeles, USA - February 21st, 2023}
}

\section{Introduction}

Children are increasingly turning to search engines to seek information on the Internet \cite{Azpiazu2017-fx,Rowlands2008-uo,pilgrim2019we,gwizdka2017analysis,usta2014k}, but these young children (ages 6-11) are still in the process of learning literacy skills which, as argued in \cite{Madrazo_Azpiazu2018-pn}, affects how they search for and consume information. Web search for children seems like an obvious setting where an intelligent conversational agent such as a spoken dialogue system or interactive virtual agent could make the search process easier by automatically transcribing child speech into search terms (so the child doesn't have to type), performing the search, then selecting and reading aloud the information that the agent thinks that the child is searching for. Furthermore, web search often requires multiple iterations of query formulation and an agent could help with the steps of fine-tuning the search parameters. Such an agent could enable children to more easily access web resources for entertainment as well as educational purposes (see \cite{Garg2022-jp} for a review of children and voice-based conversational agents). 

Following recent work \cite{Aliannejadi2022-op} that explored how technology can sometimes hinder child learning, in this paper, we take the position that an automated agent should act as a facilitator in the search process, not actually perform the work of web search (including transcription) for the child because doing so can potentially hinder the development of critical literacy and technical skills. We explain some of the functionality such an agent should have and the methodology for designing the agent. Moreover, we take the stance that the agent should be perceived as an adult taking on a teacher-facilitator role --- not as a child peer. Though web search seems like a relatively narrow use case for automated agents that interact with children, search is a common and important setting that children begin using early and we can derive an important guiding principle for all child-directed conversational agent research: agents should not do for children what children can do, or should learn to do, for themselves. 


\section{Web Search: A Setting for Learning Critical Skills}

Web search is an important domain to consider because many children use web search tools almost daily \cite{Dragovic2016-rt,gossen2013specifics} and search is a setting where children's writing and reading skills enable them to interact with a broader and increasingly important part of their digital world. Effective web search requires the application of multiple skills that are required for effective query formulation, including literacy (i.e., typing, spelling words, composing words into phrases), understanding how search results are structured,  determining if a resource returned by a search engine is relevant (i.e., what they were looking for, and reformulating their query to refine their search). This follows recent work in calling for a \emph{search as learning} paradigm \cite{gwizdka2016search} that focuses on what web search is good for: finding information and learning from a wealth of public resources, but doing so requires properly training children in classroom settings \cite{Landoni2022-fw}.

Some of the above-listed skills required for web search are often skipped altogether when using a conversational agent that automatically transcribes a verbal request and replies with a specific answer instead of a listing of resources. While this pattern of interaction allows children to access information more directly in a context of developing their literacy skills, one cannot assume that the agent's response actually fulfills the child's particular search need \cite{yarosh2018children,lovato2019hey}. Even children who like using Amazon Echo to request information often do not trust the information they are given \cite{Wojcik2022-fd}, which highlights yet another skill that children need to acquire besides literacy and search: a healthy skepticism that technology always has the correct answer \cite{murray2021more}. A traditional graphical web search interface is not so direct: they don't just give the answer; rather, they lead a child searcher to potentially relevant resources, but the child has to determine if the resource fulfills their search need. 

Our position, therefore, is that conversational agents can help facilitate children to accomplish tasks, but those agents should not do what children need to learn to do for themselves, but can provide scaffolding to help children learn.

\section{Agent Facilitator: Requirements and Methods for Development}

Here we sketch some of the requirements for a child-directed conversational agent that facilitates learning about web search. The first requirement is that the agent should be viewed by the child as an adult teacher.\footnote{This is not to say that child agents are never useful; see for example \cite{Vogt2017-gn} that explored second language tutoring with a child-like agent, but researchers should take into consideration the goals and outcomes of the tutoring when deciding about agent morphology.} In \cite{Downs2020-gf} we showed that children prefer adult voices in correcting spelling mistakes, a small yet crucial aspect of the search process. Moreover, \cite{Landoni2020-nz} showed that children do not prefer a virtual search agent to be a peer. This suggests that when corrections take place, it is better for the agent to have a clear role of being a teacher-like facilitator.  

\cite{Landoni2020-nz} explored child-directed (ages 9-11) search agents by asking the children to illustrate ``Sonny," a fictitious virtual agent that could help them find information, then the researchers asked the children what they thought Sonny should be able to do from a list of options. While the children indicated that they wanted the agent to be able to talk (i.e., converse verbally) with them, the children also indicated that they wanted to feel safe, have fun, and that the agent should remember their previous requests and take care of privacy. We point out here that the children may have identified memory as an important aspect of the search because speech-based assistants like Alexa do not remember dialogue context beyond the current request, making the ability (or lack thereof) salient to the children. Remembering prior interactions is crucial to not only to learn about a particular child's preferences, but, perhaps more importantly, to recognize the literary and search skill levels of the child and how to help the child improve those skills. 

The role of the agent should be clear to the child, and here we are advocating for a teacher-facilitator role. An agent that takes on a facilitator role should use proper pedagogical methodology because children are in a crucial developmental stage where they are learning many new things, and learning best occurs when children are given opportunities to attempt challenging tasks with proper support. If the task is too easy (i.e., the agent completes the task for the child), the child will not learn. If the task is too hard and the child is not provided scaffolded support that will help them be successful, they will become frustrated and disengaged. A facilitator agent can ensure that children are working within this ``Zone of Proximal Development" (ZPD) \cite{vygotsky1978mind} -- where a task is challenging enough for them to learn while not causing excessive frustration.

To illustrate how ZPD would work in a web search setting, consider an agent that recognizes a misspelling and flatly utters \emph{that's spelled incorrectly}. This would not help a child improve their spelling and could negatively reinforce to the child that making mistakes is bad. In this instance, the child will not improve as a speller or searcher. However, if the agent simply corrects the spelling automatically without drawing any attention to the error, the child is not provided any opportunity to learn. Rather, an inquisitive attitude of \emph{hmm, the spelling there doesn't look right, can you check the spelling of that word?}, coupled with spelling suggestions, gives the impression that being willing to make mistakes and correct them is part of the learning process. It also gives the child an opportunity to review and identify the spelling error on their own, rather than the error being auto-corrected by the system. This means that not only should the agent take on a facilitator role, but as argued in \cite{Murgia2021-xe,Fails2022-xr}, teachers should be part of the design process of how the agents look and how the agents act. 

Recent work that focused on an ``Effective reading partner" conversational agent did not just read to the child; the agent asked questions and provided feedback on a small set of topics relating to weather. The authors claim that the agent could be used to enhance the motivation for children to read \cite{Xu2020-gz}, likely because the agent is mirroring what a good teacher does during a classroom read-aloud. Skilled teachers build engagement around a text by asking students questions, thinking-aloud, re-phrasing questions, and building connections between the text and the world. Oftentimes algorithms and models are trained to provide a specific result or classification; teachers are well-trained to scaffold and support children as they learn and apply skills (which no agent will fully replace). In some cases, teachers may answer questions directly (as Google sometimes does at the top of its search results page), but teachers also help children think about what the answers mean. Conversational agents for children need to build on both kinds of training. In order to best capture the wealth of knowledge teachers have and to better address children's particular needs, we also advocate for a participatory design approach where diverse stake-holders can utilize their experiences, perspectives, and expertise to collaboratively co-construct a solution. The scaffolding required to effectively implement a teacher-facilitator conversational agent requires that teacher's expertise be well represented. Additionally, children -- as the intended users of the system -- can also lend valuable insights into the design of technology as they co-design with adults \cite{fails_methods_2013}.


\section{Conclusion: Better Building Together}


In this paper, we present a case for teacher-facilitator agents in the domain of information discovery. The agent should have some adult-like anthropomorphic characteristics that clearly signal their role to a child (e.g., adult voice) and the agent should operate in the Zone of Proximal Development to ensure that the child is stretched, but not overwhelmed as the child learns how literary and technical skills complement each other. We therefore advocate that teachers who have training and experience in educating children should be part of the design process of conversational agents that interact with children, and we also believe that children should have a voice in the design process.

\begin{acknowledgement}
Thank you to the anonymous reviewers for their helpful feedback. Work funded by NSF Award \# 1763649.
\end{acknowledgement}

\bibliographystyle{unsrt}
\bibliography{paperpile,refs}

\begin{thebibliography}{10}

\bibitem{Azpiazu2017-fx}
Ion~Madrazo Azpiazu, Nevena Dragovic, Maria~Soledad Pera, and Jerry~Alan Fails.
\newblock Online searching and learning: {YUM} and other search tools for
  children and teachers.
\newblock {\em Inf. Retr. Boston.}, 20(5):524--545, October 2017.

\bibitem{Rowlands2008-uo}
Ian Rowlands, David Nicholas, Peter Williams, Paul Huntington, Maggie
  Fieldhouse, Barrie Gunter, Richard Withey, Hamid~R Jamali, Tom Dobrowolski,
  and Carol Tenopir.
\newblock The google generation: the information behaviour of the researcher of
  the future.
\newblock {\em Aslib Proc.}, 60(4):290--310, January 2008.

\bibitem{pilgrim2019we}
Jodi Pilgrim.
\newblock Are we preparing students for the web in the wild? an analysis of
  features of websites for children.
\newblock {\em The Journal of Literacy and Technology}, 20(2):97--124, 2019.

\bibitem{gwizdka2017analysis}
Jacek Gwizdka and Dania Bilal.
\newblock Analysis of children's queries and click behavior on ranked results
  and their thought processes in google search.
\newblock In {\em Proceedings of the 2017 conference on conference human
  information interaction and retrieval}, pages 377--380, 2017.

\bibitem{usta2014k}
Arif Usta, Ismail~Sengor Altingovde, Ibrahim~Bahattin Vidinli, Rifat Ozcan, and
  {\"O}zg{\"u}r Ulusoy.
\newblock How k-12 students search for learning? analysis of an educational
  search engine log.
\newblock In {\em Proceedings of the 37th international ACM SIGIR conference on
  Research \& development in information retrieval}, pages 1151--1154, 2014.

\bibitem{Madrazo_Azpiazu2018-pn}
Ion Madrazo~Azpiazu, Nevena Dragovic, Oghenemaro Anuyah, and Maria~Soledad
  Pera.
\newblock Looking for the movie seven or sven from the movie frozen? a
  multi-perspective strategy for recommending queries for children.
\newblock In {\em Proceedings of the 2018 Conference on Human Information
  Interaction \& Retrieval}, CHIIR '18, pages 92--101, New York, NY, USA, March
  2018. Association for Computing Machinery.

\bibitem{Garg2022-jp}
Radhika Garg, Hua Cui, Spencer Seligson, Bo~Zhang, Martin Porcheron, Leigh
  Clark, Benjamin~R Cowan, and Erin Beneteau.
\newblock The last decade of {HCI} research on children and voice-based
  conversational agents.
\newblock In {\em Proceedings of the 2022 {CHI} Conference on Human Factors in
  Computing Systems}, number Article 149 in CHI '22, pages 1--19, New York, NY,
  USA, April 2022. Association for Computing Machinery.

\bibitem{Aliannejadi2022-op}
Mohammad Aliannejadi, Theo Huibers, Monica Landoni, Emiliana Murgia, and
  Maria~Soledad Pera.
\newblock The effect of prolonged exposure to online education on a classroom
  search companion.
\newblock In {\em Experimental {IR} Meets Multilinguality, Multimodality, and
  Interaction}, pages 62--78. Springer International Publishing, 2022.

\bibitem{Dragovic2016-rt}
Nevena Dragovic, Ion Madrazo~Azpiazu, and Maria~Soledad Pera.
\newblock ``is sven seven?'': A search intent module for children.
\newblock In {\em Proceedings of the 39th International {ACM} {SIGIR}
  conference on Research and Development in Information Retrieval}, SIGIR '16,
  pages 885--888, New York, NY, USA, July 2016. Association for Computing
  Machinery.

\bibitem{gossen2013specifics}
Tatiana Gossen and Andreas N{\"u}rnberger.
\newblock Specifics of information retrieval for young users: A survey.
\newblock {\em Information Processing \& Management}, 49(4):739--756, 2013.

\bibitem{gwizdka2016search}
Jacek Gwizdka, Preben Hansen, Claudia Hauff, Jiyin He, and Noriko Kando.
\newblock Search as learning (sal) workshop 2016.
\newblock In {\em Proceedings of the 39th International ACM SIGIR conference on
  Research and Development in Information Retrieval}, pages 1249--1250, 2016.

\bibitem{Landoni2022-fw}
Monica Landoni, Maria~Soledad Pera, Emiliana Murgia, and Theo Huibers.
\newblock Let's learn from children: Scaffolding to enable search as learning
  in the educational environment.
\newblock September 2022.

\bibitem{yarosh2018children}
Svetlana Yarosh, Stryker Thompson, Kathleen Watson, Alice Chase, Ashwin
  Senthilkumar, Ye~Yuan, and AJ~Bernheim Brush.
\newblock Children asking questions: speech interface reformulations and
  personification preferences.
\newblock In {\em Proceedings of the 17th ACM conference on interaction design
  and children}, pages 300--312, 2018.

\bibitem{lovato2019hey}
Silvia~B Lovato, Anne~Marie Piper, and Ellen~A Wartella.
\newblock Hey google, do unicorns exist? conversational agents as a path to
  answers to children's questions.
\newblock In {\em Proceedings of the 18th ACM international conference on
  interaction design and children}, pages 301--313, 2019.

\bibitem{Wojcik2022-fd}
Erica~H Wojcik, Aarathi Prasad, Samantha~P Hutchinson, and Kyla Shen.
\newblock Children prefer to learn from smart devices, but do not trust them
  more than humans.
\newblock {\em International Journal of Child-Computer Interaction}, 32:100406,
  June 2022.

\bibitem{murray2021more}
Grace~W Murray.
\newblock Who is more trustworthy, alexa or mom?: Children’s selective trust
  in a digital age.
\newblock 2021.

\bibitem{Vogt2017-gn}
Paul Vogt, Mirjam de~Haas, Chiara de~Jong, Peta Baxter, and Emiel Krahmer.
\newblock Child-robot interactions for second language tutoring to preschool
  children.
\newblock {\em Front. Hum. Neurosci.}, 11:73, March 2017.

\bibitem{Downs2020-gf}
Brody Downs, Aprajita Shukla, Mikey Krentz, Maria~Soledad Pera,
  Katherine~Landau Wright, Casey Kennington, and Jerry Fails.
\newblock Guiding the selection of child spellchecker suggestions using audio
  and visual cues.
\newblock In {\em Proceedings of the Interaction Design and Children
  Conference}, IDC '20, pages 398--408, New York, NY, USA, June 2020.
  Association for Computing Machinery.

\bibitem{Landoni2020-nz}
Monica Landoni, Emiliana Murgia, Theo Huibers, and Maria~Soledad Pera.
\newblock You've got a friend in me: Children and search agents.
\newblock In {\em Adjunct Publication of the 28th {ACM} Conference on User
  Modeling, Adaptation and Personalization}, UMAP '20 Adjunct, pages 89--94,
  New York, NY, USA, July 2020. Association for Computing Machinery.

\bibitem{vygotsky1978mind}
Lev~Semenovich Vygotsky and Michael Cole.
\newblock {\em Mind in society: Development of higher psychological processes}.
\newblock Harvard university press, 1978.

\bibitem{Murgia2021-xe}
Emiliana Murgia, Monica Landoni, Theo Huibers, and Maria~Soledad Pera.
\newblock All together now: Teachers as research partners in the design of
  search technology for the classroom.
\newblock May 2021.

\bibitem{Fails2022-xr}
Jerry~Alan Fails, Monica Landoni, Theo Huibers, and Maria~Soledad Pera.
\newblock Report on the 5th workshop on international and interdisciplinary
  perspectives on children \& recommender and information retrieval systems
  ({KidRec} 2021) at {IDC} 2021: the teacher lens.
\newblock {\em SIGIR Forum}, 55(2):1--6, March 2022.

\bibitem{Xu2020-gz}
Ying Xu and Mark Warschauer.
\newblock Exploring young children's engagement in joint reading with a
  conversational agent.
\newblock In {\em Proceedings of the Interaction Design and Children
  Conference}, IDC '20, pages 216--228, New York, NY, USA, June 2020.
  Association for Computing Machinery.

\bibitem{fails_methods_2013}
Jerry~Alan Fails, Mona~Leigh Guha, and Allison Druin.
\newblock {\em Methods and {Techniques} for {Involving} {Children} in the
  {Design} of {New} {Technology} for {Children}}.
\newblock Now Publishers Inc., Hanover, MA, USA, 2013.

\end{thebibliography}

\end{document}